\documentclass[pdflatex,sn-basic,iicol]{sn-jnl}

\usepackage{graphicx}%
\usepackage{multirow}%
\usepackage{amsmath,amssymb,amsfonts}%
\usepackage{amsthm}%
\usepackage{mathrsfs}%
\usepackage[title]{appendix}%
\usepackage{xcolor}%
\usepackage{textcomp}%
\usepackage{manyfoot}%
\usepackage{booktabs}%
\usepackage{algorithm}%
\usepackage{algorithmicx}%
\usepackage{algpseudocode}%
\usepackage{listings}%
\usepackage{arydshln}

\theoremstyle{thmstyleone}%
\theoremstyle{thmstyletwo}%

\theoremstyle{thmstylethree}%

\graphicspath{{./figures/}}

\usepackage{xcolor}
\usepackage{tikz}
\definecolor{gold}{HTML}{BD820B}
\definecolor{silver}{HTML}{909090}
\definecolor{bronze}{HTML}{9A5F26}

\newcommand*\circledd[1]{\tikz[baseline=(char.base)]{
            \node[shape=circle,draw,inner sep=0.15pt] (char) {#1};}}     
            
\newcommand{\first}[1]{%
    {#1\raisebox{0.8pt}{\footnotesize \color{gold} \circledd{1}}}%
}
\newcommand{\second}[1]{%
    {#1\raisebox{0.8pt}{\footnotesize \color{silver} \circledd{2}}}%
}
\newcommand{\third}[1]{%
    {#1\raisebox{0.8pt}{\footnotesize \color{bronze} \circledd{3}}}%
}

\begin{document}

\title[Distractor-Aware Memory-Based Visual Object Tracking]{Distractor-Aware Memory-Based Visual Object Tracking}

\author*[1]{\fnm{Jovana} \sur{Videnović}}\email{jovanavidenovic10@gmail.com}

\author[1]{\fnm{Matej} \sur{Kristan}}\email{matej.kristan@fri.uni-lj.si}

\author[1]{\fnm{Alan} \sur{Lukežič}}\email{alan.lukezic@fri.uni-lj.si}

\affil[1]{\orgdiv{Faculty of Computer and Information Science}, \orgname{University of Ljubljana}, \orgaddress{\street{Večna pot 113}, \city{Ljubljana}, \postcode{1000}, \country{Slovenia}}}

\abstract{
Recent emergence of memory-based video segmentation methods such as SAM2
has led to models with excellent performance in segmentation tasks, achieving leading results on numerous benchmarks. 
However, these modes are not fully adjusted for visual object tracking, where distractors (i.e., objects visually similar to the target) pose a key challenge.
In this paper we propose a distractor-aware drop-in memory module and introspection-based management method for SAM2, leading to DAM4SAM.
Our design effectively reduces the tracking drift toward distractors and improves redetection capability after object occlusion. 
To facilitate the analysis of tracking in the presence of distractors, we construct DiDi, a Distractor-Distilled dataset. DAM4SAM outperforms SAM2.1 on thirteen benchmarks and sets new state-of-the-art results on ten. 
Furthermore, integrating the proposed distractor-aware memory into 
a real-time tracker EfficientTAM 
leads to 11\% improvement 
and matches tracking quality of the non-real-time SAM2.1-L on multiple tracking and segmentation benchmarks, while integration with edge-based tracker EdgeTAM delivers 4\% performance boost, demonstrating a very good generalization across architectures. The code for this work and the new distractor-distilled dataset are available at \url{https://github.com/jovanavidenovic/DAM4SAM}.
}


\maketitle

\section{Introduction}\label{sec1}
\begin{figure*}[t]
  \centering
  \includegraphics[width=\linewidth]{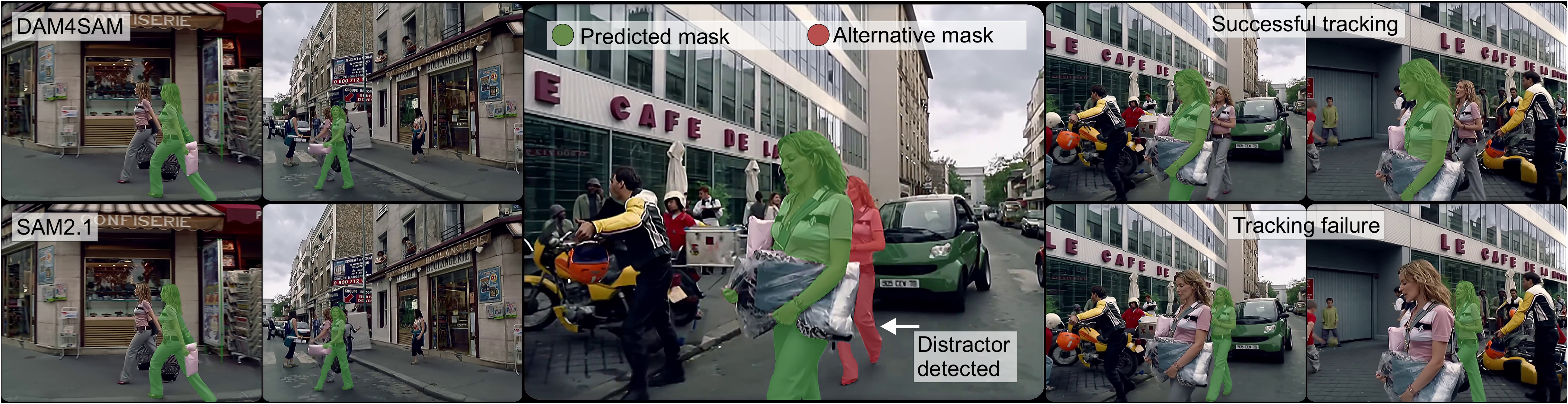} 
\caption{
Comparison of DAM4SAM (top row) with SAM2.1 (bottom row). The proposed distractor-aware memory (DAM), integrated into SAM2.1 (referred to as DAM4SAM), resolves visual ambiguities and enhances tracking robustness. The DAM update is triggered by divergence between the predicted and alternative masks in SAM2.1 multi-mask output (shown in the middle column). At this point, both SAM2.1 and DAM4SAM still track the target correctly, but one of the alternative mask segments the distractor, which triggers the DAM update, leading to improved performance compared to SAM2.1 (last two columns).}
\label{fig:fig1}
\end{figure*}

General visual object tracking is a core task in computer vision that involves localizing an arbitrary target across a video, by a bounding box or a segmentation mask, given a single example in the first frame. Over the past decade, the significant progress has been driven by deep learning and transformer architectures~\citep{transt_cvpr2021, ms_aot, lorat_eccv2024}, the availability of diverse training datasets~\citep{lasot_cvpr19,got10k, muller_trackingnet}, and the development of challenging benchmarks~\citep{otb_pami2015, vot2022, vots2024, lasot_ijcv}. Modern trackers are now widely used in applications such as surveillance, video editing, and sports analytics.

Nevertheless, a substantial source of tracking failures remain nearby objects that visually resemble the tracked object. These regions, called distractors, increase the localization uncertainty and mislead the tracker (Figure~\ref{fig:fig1}). Distractors can be classified as external -- nearby objects similar to the target -- or internal -- self-similar patterns within the target itself when only a part is being tracked. 
While internal distractors mainly reduce localization accuracy, external distractors present a particular challenge when the target temporarily leaves and re-enters the frame, frequently causing irreversible tracking drift.

Various approaches have been proposed to reduce the visual ambiguity caused by distractors. 
These include learning discriminative features~\citep{siamfc_eccvw2016,stark_iccv21,transt_cvpr2021,mixformer_cvpr2022, seqtrack} or explicitly modeling the foreground-background by dedicated modules~\citep{d3s_tpami,odtrack,cutie_cvpr2024,ditra_ijcv2024}.
An emerging paradigm, already positioned at the top of the major benchmarks~\citep{vot2022, vots2023, vots2024}, are memory-based frameworks, which localize the target by pixel association with the past tracked frames~\citep{ms_aot,xmem_eccv2022,rmem_cvpr2024}.

Memory-based methods construct the target model by storing multiple past images with the target segmented by the tracker, thereby implicitly encoding the potential distractors segmented as the background. While the tendency of initial memory-based models was to continually add arriving frames to the memory, ~\cite{rmem_cvpr2024} argued that the visual redundancy in large memory leads to reduced localization capability due to the nature of cross-attention. They show that limiting the memory to the most recent frames and temporally time-stamping them in fact improves tracking.
This paradigm was further validated by the recent video segmentation foundation model SAM2~\citep{sam2}, which sets a solid state-of-the-art across several video segmentation and tracking benchmarks~\citep{lvos_iccv2023,ytvos,  mose,vots2024, davis17}. 

We argue that while the recent target appearances in the memory are required for accurate segmentation, another type of memory is required to distinguish the target from challenging distractors.
To support this claim, we propose a new distractor-aware memory (DAM) and a memory maintanance mechanism for SAM2.
The new memory is divided by its tracking functionality into two parts: the recent appearances memory (RAM) and distractor-resolving memory (DRM). 
While RAM contains the recent target appearances sampled at regular intervals, DRM contains anchor frames that help discriminate the target from critical distractors. 
A novel DRM updating mechanism is proposed that exploits the multiple hypothesis output of SAM2 (see Figure~\ref{fig:fig1}), which has been previously overlooked in the tracking research.

In addition, we observe that standard benchmarks contain many sequences, which are no longer considered challenging by modern standards. The high performance on these sequences overwhelms the total score, saturates the benchmarks, and does not properly expose the tracking advances. 
To address this, we semi-automatically distill several benchmarks into a distractor-distilled tracking dataset (DiDi). The dataset contains 180 sequences, each annotated with the initial segmentation mask. In our experimental analysis, we show that DiDi clearly exposes the tracking improvements of trackers with explicit distractor handling compared to others, which is not the case for other common benchmarks~\citep{got10k, lasot_cvpr19, lasot_ijcv}.

In summary, we make the following contributions:
\begin{itemize}
    \item We propose a drop-in distrator-aware memory (DAM) framework for memory-based trackers. To the best of our knowledge, this is the first memory formulation that divides and updates the memory with respect to its function in tracking.
    \item We integrate DAM into SAM2.1~\citep{sam2} model, leading to DAM4SAM. Without requiring additional training, DAM4SAM improves a baseline tracker by notable margins on thirteen visual object tracking and video object segmentation benchmarks, setting state-of-the-art results on ten of them, including LaSOT$_{ext}$~\citep{lasot_ijcv} and LVOSv2~\citep{lvosv2_arxiv2024}.
    \item We integrate DAM into the real-time SAM2-based model, named Efficient Track Anything (TAM)~\citep{efficient_tam}. This demonstrates DAM’s generalization capability, as it boosts TAM's performance by up to 11\%. Notably, DAM-enhanced TAM achieves tracking accuracy comparable to SAM2.1-L on several datasets, despite using 6.5$\times$ fewer parameters and being 2$\times$ faster. 
    \item We construct a new DiDi dataset. The dataset consists of 180 sequences, with average length of 1,500 frames, making it the largest and most challengig distractor-oriented tracking dataset.
\end{itemize}

A preliminary version of this work was published in~\citep{dam4sam}.
This paper extends it in several directions: we extend the related work by 
charting historical developments in tracking that led to 
memory-based trackers and include concurrent SAM2 extensions, we expand the experimental evaluation to include six additional datasets, including three video object segmentation benchmarks, extend the ablation study and present the sensitivity analysis, demonstrate generalization capabilities of DAM, and provide detailed qualitative insights with a failure case analysis.

\section{Related work}
Early deep learning methods in visual object tracking focused on fine-tuning pre-trained networks, or their parts, during inference~\citep{mdnet_cvpr2016, atom_cvpr19}. While these approaches outperformed traditional methods such as correlation filter-based trackers~\citep{kcf_pami2014, samf_eccvw2014}, they were computationally intensive and struggled with robustness to object appearance variations. Fine-tuning allowed the model to adapt to the specific target object, but the need for online updates introduced inference latency.

A major shift occurred with the emergence of CNN-based Siamese trackers~\citep{siamfc_eccvw2016, siamrpn_cvpr2018, siamrpn_cvpr2019, siamcar_cvpr2020}, which eliminated online training by adopting a template-matching paradigm using a reference template from the first frame. These methods brought notable efficiency gains but struggled with distinguishing the target from semantically similar backgrounds. Their reliance on local search strategies also limited robustness under full occlusion or out-of-view scenarios.

The issues of CNN-based Siamese trackers were partially addressed by attention-based models~\citep{transt_cvpr2021, simtrack_eccv2022}, which extended the Siamese architecture by employing attention mechanisms to fuse template and search region features. For instance, they achieved state-of-the-art performance witout even updating the template~\citep{transt_cvpr2021}, while combining efficiency and architectural simplicity. 
The next generation of Transfomer-based trackers incorporated information beyond the initial frame for localization of the target~\citep{stark_iccv21, seqtrack}. This family of methods leveraged transformer-based architectures to dynamically update the target template using predictions from the previous frame, which enabled adaptive and temporally consistent tracking, improving robustness to occlusions, deformations, and appearance changes.

In the past years, the tracking paradigm of top-performers has shifted towards video segmentation-based memory networks~\citep{xmem_eccv2022, ms_aot,cutie_cvpr2024,sam2, rmem_cvpr2024}. These methods embed predictions from past frames into memory, therefore extending contextual information beyond just the initial or the previous frame.
The attention mechanism is typically used to link frame representations stored in the memory with features extracted from the current frame.
In initial methods such as by~\citep{ms_aot}, arriving frames were continually added to the memory. This led to theoretically unbounded increase in computational complexity and GPU memory. This issue was addressed in~\citep{xmem_eccv2022, cutie_cvpr2024} by using multiple memory storages and efficient compression schemes to capture different temporal contexts, enhancing  performance on long-term videos. 
Later,~\cite{rmem_cvpr2024} proposed to restrict the memory to the most recent frames with temporal stamping, which led to improved localization. 

Recently, ~\cite{sam2} introduced a restricted memory-based video segmentation foundation model SAM2, which demonstrated remarkable performance across a number of video object segmentation and tracking benchmarks. The model retains the initial frame and the six most recent frames in memory. Its remarkable perormance spawned a number of of follow-up works that built on the proposed architecture. ~\cite{efficient_tam} proposed EfficientTAM, a lightweight tracking model with results comparable to SAM2, higher speed and smaller model sizes. 
Building on SAM2~\citep{sam2}, several concurrent works have extended it to tackle specialized tasks, including long-term video object segmentation~\citep{sam2long}, motion-aware object tracking~\citep{samurai}, and referring segmentation in videos~\citep{samwise}. 
SAM2Long~\citep{sam2long} selects the optimal segmentation trajectory via a constrained tree search, thereby achieving better results on long sequences. 
SAMURAI~\citep{samurai} integrates motion cues into memory selection and mask refinement, which enables better handling of crowded scenes with fast motions.
These studies are concurrent and closely related to ours, as they share the core idea of adapting memory to address specific challenges. 
However, DAM4SAM differs by introducing a new memory model that separates the memory according to its function and updates it by exploiting SAM2 output ambiguity for explicit distractor detection, resulting in improved segmentation accuracy and robustness.

Several works have recognized the challenge of tracking in presence of distractors and proposed different solutions to the problem. ~\cite{keeptrack} cast the problem as a multi-target tracking setup and propose KeepTrack. The model identifies target candidates and potential distractors, which are then associated with previously propagated identities using a learned association network. However, the method relies on accurate detection and cannot address internal distractors in practice. 
~\cite{ditra_ijcv2024} showed that treating target localization accuracy and robustness as two distinct tasks is beneficial, particularly in scenarios involving distractors. 
Nevertheless, these distractor handling pipelines lead to complex pipelines, potentially combining several methods with many hand-crafted rules, thus preventing them from fully exploiting the learning potential of modern frameworks. 
Alternatively, memory-based methods~\citep{xmem_eccv2022, cutie_cvpr2024, rmem_cvpr2024, sam2} have the capacity to implicitly handle distractors, since they store entire images and apply a learnable localization by segmentation.
We make a step further in DAM4SAM by proposing to split the memory according to its function of distractor learning and frame-to-frame segmentation, and developing a novel memory update technique based on distractor detection.

\section{DAM4SAM}\label{sec:method}
This section describes the new distractor-aware memory (DAM) model for SAM2, which is a drop-in module for SAM2 and does not require additional training. 
For completeness, the SAM2 architecture is described in Section~\ref{sec:sam2}, while the new model is described in Section~\ref{sec:memory_management}.

\subsection{SAM2 preliminaries}  \label{sec:sam2}

SAM2 extends the Segment Anything Model (SAM)~\citep{sam_iccv2023}, originally developed for interactive class-agnostic image segmentation, to video segmentation. It consists of four main components: (i) image encoder, (ii) prompt encoder, (iii) memory bank, and (iv) mask decoder.

The image encoder applies the Hiera\footnote{Hiera-L version is used in all our experiments.} 
backbone~\citep{hiera} to embed the input image.
Interactive inputs (e.g., positive/negative clicks) are absorbed by the prompt encoder and used for output mask refinement. 
Note that in the non-interactive general object tracking setup considered in this paper, these prompts are used only during the initialization step and are not applied during tracking. 
The memory bank consists of the encoded initialization frame with a user-provided segmentation mask and six recent frames with segmentation masks generated by the tracker. Temporal encodings are applied to the six recent frames to encode the frame order, while such encoding is not applied to the initialization frame to indicate its unique property of being a single supervised training example and thus serves as a sort of target prior model.

The memory bank transfers pixel-wise labels onto the current image by cross-attending the features in the current frame to all memory frames, producing memory-conditioned features. The features are then decoded by the mask decoder, which predicts three output masks along with the prediction of the IoU with the (unknown) ground truth for each. The mask with the highest predicted IoU score is chosen as the tracking output.

SAM2 applies a variant of the memory management proposed in~\cite{rmem_cvpr2024}, where the initialization frame is always kept in the memory, while the six recent frames are updated at every new frame by a first-in-first-out protocol. The memory and the management mechanism are visualized in  Figure~\ref{fig:memory}, while the reader is referred to~\cite{sam2} for more details.

\subsection{Distractor-aware memory}  \label{sec:memory_management}

Related works~\citep{ms_aot,xmem_eccv2022,rmem_cvpr2024,sam2} have clearly demonstrated the importance of the most recent frames to address target appearance changes and ensure accurate segmentation. However, a different type of frames is required to prevent drifting in the presence of critical distractors and for reliable target re-detection.

\begin{figure*}[h]
  \centering
  \includegraphics[width=\linewidth]{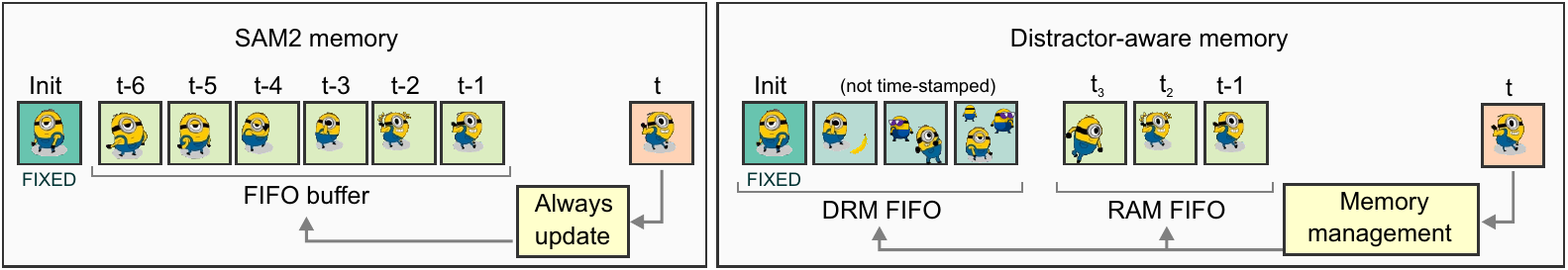} 
    \caption{Overview of the SAM2 memory and the proposed Distractor-Aware Memory (DAM), which splits the model into {\it Recent Appearance Memory} (RAM) and {\it Distractor Resolving Memory} (DRM), and updates them using a new memory management protocol. 
    Note that $t_2$ and $t_3$ represent the time steps at which RAM was updated, i.e., every $\Delta$ frames. Additionally, RAM and DRM update is prevented on frames where the predicted mask is empty.}
    \label{fig:memory}
\vspace{-0.2cm}
\end{figure*}

We propose to separate the memory with respect to its function during tracking into (i) \textit{recent appearance memory} (RAM) and (ii) \textit{distractor resolving memory} (DRM). 
RAM and DRM together form the distractor-aware memory (DAM), visualized in Figure~\ref{fig:memory}. 
The function of RAM is to ensure segmentation accuracy in the considered frame. 
Thus we design it akin to the current SAM2~\citep{sam2} memory, as a FIFO buffer with 
${1 \over 2}N_\mathrm{DAM}=3$
slots, containing the most recent target appearances with temporal encoding to identify temporally more relevant frames for the task. 

On the other hand, DRM serves for ensuring tracking robustness and re-detection. It should contain accurately segmented frames with critical recent distractors, including the initialization frame. 
It is thus composed of a slot reserved for the initialization frame and a FIFO buffer with ${1 \over 2}N_\mathrm{DAM}=3$ anchor frames updated during 
tracking\footnote{At tracker initialization, RAM occupies all $N_\mathrm{DAM}$ slots and drops to ${1 \over 2}N_\mathrm{DAM}$ as DRM entries arrive, to fully exploit the available capacity.}.
Since the purpose of DRM is to encode critical information for resolving distractors, it does not apply temporal encoding. Note that the pre-trained SAM2 already contains the building blocks to implement the proposed memory structure. 

\subsubsection{RAM management protocol}  \label{sec:memory_ram}

A crucial element of the memory-based methods is the memory management protocol. 
To efficiently exploit the available memory slots, the memory should not be updated at every frame, since the consecutive frames are highly correlated. 
Indeed,~\cite{rmem_cvpr2024} argue that visual redundancy in memory should be avoided in attention-based localization. When the memory contains many images with redundant and noisy features, the attention becomes diluted, which can lead to lower weights on the truly relevant features. In visual object tracking, the videos are typically recorded at 30 FPS, resulting in high appearance redundancy between consecutive frames. RAM is thus updated every $\Delta=5$ frames and includes the most recent frame since it is the most relevant for accurate target segmentation in the considered frame.

SAM2~\citep{sam2} updates the memory at every frame, including when the target is absent. The target is considered absent when the segmentation mask is all zeros. However, even for a very short occlusion, the memory will quickly fill up with frames without the target, which reduces the target appearance diversity in the model, leading to a reduced segmentation accuracy upon target re-appearance. Furthermore, failing to re-detect the target leads to incorrectly updating the memory by an empty mask, which may cause error accumulation and ultimately re-detection failure. We thus propose to not update RAM when the target is not present, i.e., when the predicted target mask is empty. 

\subsubsection{DRM management protocol}  \label{sec:memory_drm}

DRM inherits the initial update rules from RAM, i.e., update only when the target is present and at least $\Delta=5$ frames have passed since the last update.
It considers an additional rule to identify anchor frames containing critical distractors. 
In particular, drifting to a distractor may be avoided by including a past temporally nearby frame with this distractor accurately segmented as the background. 
Recall that SAM2 predicts three output masks and selects the one with the highest predicted IoU (Section~\ref{sec:sam2}), which means we can consider it as a multi-hypothesis prediction model.
Our preliminary study showed that in the frames before the failure occurs, SAM2 in fact detects such distractors in the alternative two predicted output masks (see Figure~\ref{fig:fig1}). 
We thus propose a simple anchor frame detection mechanism based on determining hypothesis divergence between the output and alternative masks.

\begin{figure*}[h]
  \centering
  \includegraphics[width=\linewidth]{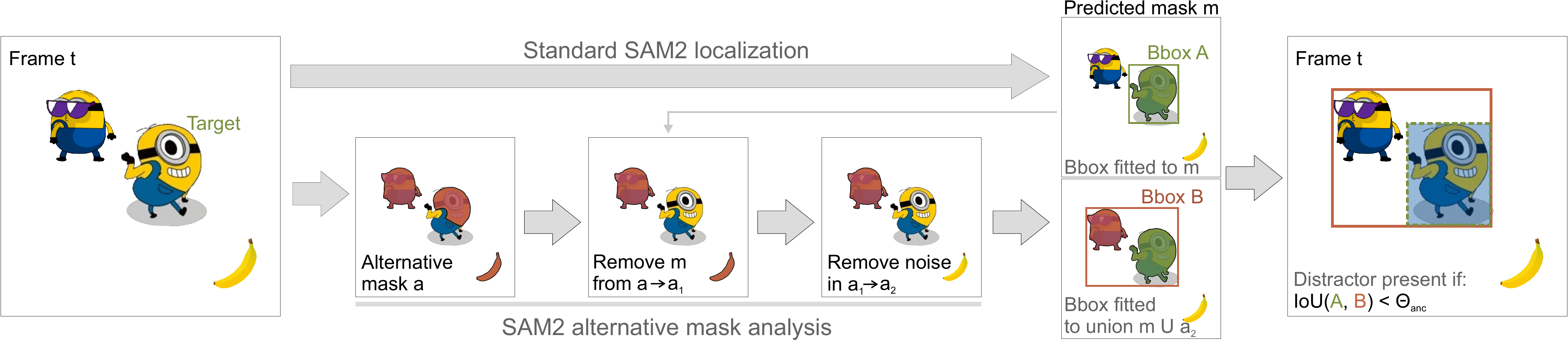} 
  \caption{Anchor frame is identified by detecting a distractor through analysis of agreement between the SAM2 predicted object mask with the predicted alternative masks.}
\label{fig:drm-update}
\end{figure*}

A bounding box is fitted to the output mask and to the union of the output mask and the largest connected component in the alternative mask. 
If the ratio between the area of the two bounding boxes drops below $\theta_\mathrm{anc}=0.7$, the current frame is considered as a potential candidate to update DRM. 
The anchor frame detection is visualized Figure~\ref{fig:drm-update}. 
Note that updating with a grossly incorrectly segmented target would lead to memory corruption and eventual tracking failure. 
We thus trigger the DRM update only during sufficiently stable tracking periods, i.e., if the predicted IoU score from SAM2 exceeds a threshold $\theta_\mathrm{IoU}=0.8$ and if the mask area is within $\theta_\mathrm{area}=0.2$ of the median area in the last $N_{M}=10$ frames. 
Note that DAM4SAM is mostly insensitive to the exact value of these parameters, as presented later in Section~\ref{sec:sensitivity_analysis}. 

\section{A distractor-distilled dataset}  \label{sec:disd_dataset}

While benchmarks played a major role in the recent visual object tracking breakthroughs, we note that many of these contain sequences, which are no longer considered challenging by modern standards. In fact, most of the modern trackers obtain high performance on these sequences, which overwhelms the total score and under-represents the improvements on challenging situations.
To facilitate the tracking performance analysis of the designs proposed in this paper, we semi-automatically distill several benchmarks into a distractor-distilled tracking dataset (DiDi).

We considered validation and test sequences of the major tracking benchmarks, which are known for high-quality annotation, i.e., GoT-10k~\citep{got10k}, LaSOT~\citep{lasot_cvpr19}, UTB180~\citep{utb180}, VOT-ST2020 and VOT-LT2020~\citep{kristan_vot2020}, and VOT-ST2022 and VOT-LT2022~\citep{vot2022}. This gave us a pool of 808 sequences. 
A sequence was selected for the DiDi dataset if at least one-third of the frames passed the distractor presence criterion described next.

A frame was classified as containing non-negligible distractors if it contained a large enough region, visually similar to the target. To make this criterion as much independent as possible from the modern tracking localization methods, we considered the following approach. First, dense image feature maps were extracted from the input image by a DINOv2~\citep{dino2} backbone. For each pixel in this feature space, we computed its cosine similarity to multiple target feature vectors extracted from the ground truth target region. The distractor score at each pixel was then calculated as the average cosine similarity between that pixel’s feature vector and the set of target features, resulting in a dense similarity response map. We then thresholded the map to obtain a binary map containing regions similar to the target. To assess the presence of distractors, we computed the ratio of the number of high-similarity pixels outside the target region to those inside. If this ratio exceeded 0.5, the frame was classified as containing distractors.

\begin{table}[h]
    \centering
    \caption{Summary of datasets used for constructing the proposed DiDi dataset.}
    \label{tab:didi_summary}
    \begin{tabular}{lc}
        \hline
        Dataset & Num. of seqs. \\
        \hline
        LaSoT~\citep{lasot_cvpr19} 
        & 86 \\
        UTB-180~\citep{utb180} & 56 \\
        VOT2022-ST~\citep{vot2022} & 20 \\
        VOT2022-LT~\citep{vot2022} & 7 \\
        VOT2020-LT~\citep{kristan_vot2020} & 6 \\
        GoT10k~\citep{got10k} & 4 \\
        VOT2020-ST~\citep{kristan_vot2020} & 1 \\
        \hline
        Total & 180 \\
        \hline
    \end{tabular}
\end{table}

\begin{figure}[ht]
  \centering
  \includegraphics[width=\linewidth]{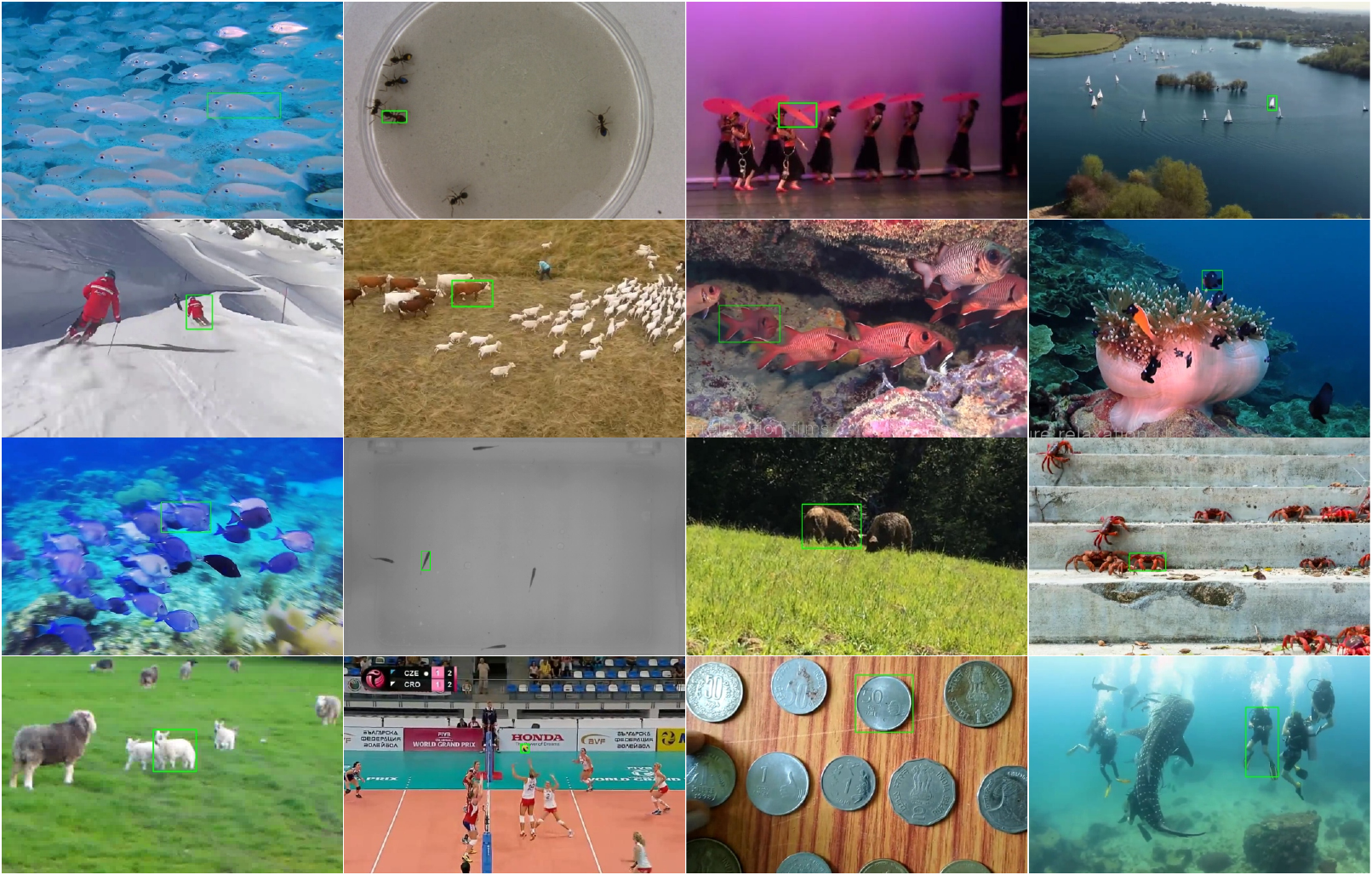} 
  \caption{Example frames from the DiDi dataset showing challenging distractors. Targets are denoted by green bounding boxes.}
\label{fig:dataset}
\end{figure}

Using the aforementioned distractor presence detection, we finally obtained 180 sequences with an average sequence length of 1,500 frames (274,882 frames in total). Each sequence contains a single target annotated by an axis-aligned bounding box. In addition, we manually segmented the initial frames to allow the initialization of segmentation-based trackers. The number of sequences obtained from each source dataset is given in Table~\ref{tab:didi_summary}. Figure~\ref{fig:dataset} shows frames from the proposed DiDi dataset. 

\section{Experiments}  \label{sec:experiments}

Various aspects of the proposed DAM for SAM2 are analyzed in the following. Section~\ref{sec:ablation} reports an ablation study -- it includes series of experiments to justify the design choices and to demonstrate sensitivity to threshold values. 
Section~\ref{sec:exp-disd} compares the SAM2.1 extension with DAM with the state-of-the-art on the DiDi dataset. 
Detailed analysis on the challenging VOT tracking-by-segmentation benchmarks is reported in Section~\ref{sec:exp-vot}. 
Comparisons on the standard bounding-box tracking and video object segmentation benchmarks are reported in Section~\ref{sec:exp-bb} and Section~\ref{sec:exp-vos}, respectively. 
In the final two Section~\ref{sec:qualitative} and Section~\ref{sec:qa_failures}, we present qualitative analysis of the proposed tracker and an overview of its failure cases. 

\subsection{Ablation study}  \label{sec:ablation}

\subsubsection{Architecture design decisions}  \label{sec:exp-justification}

\begin{figure}[ht]
  \centering
  \includegraphics[width=\linewidth]{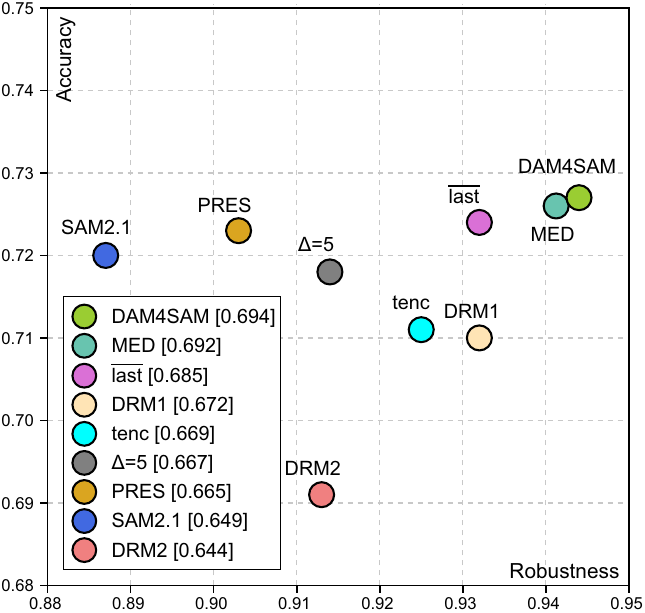} 
  \caption{
   Accuracy-robustness plot on DiDi for the ablated versions of DAM4SAM. The tracking quality is given at each label. In interest of clarity, shortened labels correspond to the subscript components of the full model names in Table~\ref{tab:architecture_justification} are used.
  }
\label{fig:ar_ablation}
\end{figure}

The design choices of the proposed distractor-aware memory and the management protocol from Section~\ref{sec:memory_management} are validated on the DiDi dataset from Section~\ref{sec:disd_dataset}. 
We apply the standard VOTS~\citep{vots2023} performance evaluation protocol and its metrics since they simultaneously account for short-term as well as long-term tracking performance. 
Performance is summarized by the tracking quality Q score and two auxiliary measures: robustness (i.e., portion of successfully tracked frames) and accuracy (i.e., the average IoU between the prediction and ground truth during successful tracking). 
Results are shown in Table~\ref{tab:architecture_justification} and in the AR plot in Figure~\ref{fig:ar_ablation}.

We first verify the argument that updating with frames without a target present causes memory degradation. 
We thus extend SAM2.1, with updating only when the predicted mask is not empty (denoted as SAM2.1$_{\mathrm{PRES}}$). 
SAM2.1$_{\mathrm{PRES}}$ increases the tracking quality Q by 2.5\%, primarily by improved robustness, which justifies our claim.

We next verify the assumption (also claimed in~\cite{rmem_cvpr2024}) that frequent updates reduce tracking robustness due to highly correlated information stored in memory. 
We reduce the update rate in SAM2.1$_{\mathrm{PRES}}$ to every 5th frame (SAM2.1$_{\Delta=5}$). 
This negligibly improves Q, but does increase the robustness by 1.2\%, which supports the claim. 
We did not observe further performance improvement with increasing $\Delta$.

Finally, we focus on our proposed distractor-resolving memory (DRM), which accounts for the tracking robustness in the presence of distractors. 
Recall that DRM is updated when a distractor is detected and under the condition that tracking is reliable -- we first test the influence of these two conditions independently. 
We thus extend SAM2.1$_{\Delta=5}$ with the new DAM and update the DRM part only during reliable tracking periods (SAM2.1$_{DRM1}$). The tracking accuracy improves slightly, while the robustness increases by 2\%. 
Alternatively, we change the rule to update only when the distractor is detected (SAM2.1$_{DRM2}$). While robustness remains unchanged with respect to SAM2.1$_{\Delta=5}$, the accuracy in fact drops. 
This is expected since distractor detection may be triggered also by the error in the target segmentation, which becomes amplified by the update. To verify this, we next apply both our proposed update DRM rules, arriving to the proposed DAM with SAM2.1 (DAM4SAM for short). 
Compared to SAM2.1$_{\Delta=5}$, we observe substantial improvement in tracking quality Q (4\%), primarily due to 3.3\% boost in robustness as well as 1.3\% boost in accuracy, taking the top-right position in the AR plot (Figure~\ref{fig:ar_ablation}) among all variants.
This conclusively verifies that DRM should be updated with distractor detected only if tracking is sufficiently reliable. 

A central element of the distractor detection algorithm from Section~\ref{sec:memory_drm} is the connected components operation, whose task is to remove the noise in the predicted mask (e.g., small spurious patches of segmented pixels away from the main target), thus stabilizing distractor detection. 
To verify the noise removal hypothesis, we replace it by a median filter with kernel size $3\times 3$ (named DAM4SAM$_{\text{MED}}$). 
The results in Table~\ref{tab:architecture_justification} indicate that the tracking performance is comparable to when using the connected components. 
The performance drop is negligible, which strongly supports our hypothesis.

\begin{table}[th]
  \centering
  \resizebox{\linewidth}{!}{\begin{minipage}{\linewidth}
\centering
\caption{DAM4SAM architecture justification on the proposed DiDi dataset.}
\label{tab:architecture_justification}
\begin{tabular}{llll}
\toprule
  & Quality & Accuracy & Robustness \\
\midrule
SAM2.1 \tiny{(ICLR25)}       & 0.649 & 0.720 & 0.887 \\
SAM2.1$_{\mathrm{PRES}}$       & 0.665 & 0.723 & 0.903 \\
SAM2.1$_{\Delta=5}$ & 0.667 & 0.718 & 0.914 \\
SAM2.1$_{\mathrm{DRM1}}$   & 0.672 & 0.710 & 0.932 \third{} \\ 
SAM2.1$_{\mathrm{DRM2}}$  & 0.644 & 0.691 & 0.913 \\ 
DAM4SAM$_{\mathrm{MED}}$ & 0.692 \second{} & 0.726 \second{} & 0.941 \second{} \\
DAM4SAM      & \textbf{0.694} \first{} & \textbf{0.727} \first{} & {\bf 0.944} \first{} \\
\midrule
DRM$_{\mathrm{tenc}}$       & 0.669 & 0.711 & 0.925  \\
RAM$_{\mathrm{\overline{last}}}$       & 0.685 \third{} & 0.724 \third{} & 0.932 \third{} \\
\bottomrule
\end{tabular}
\end{minipage}}
\end{table}

In Section~\ref{sec:memory_management} we claim that DRM part of the memory should not be time-stamped, since the frame influence on the distractor resolving in the current frame should not be biased by the temporal proximity and should serve as a time-less prior. 
To test this claim, we modify DAM4SAM by using temporal encodings in DRM (except for the initialization frame) -- we denote it as DRM$_{\mathrm{tenc}}$. 
The tracking quality drops by 3.6\%, which confirms the claim. 

We further inspect the updating regime in RAM, which always includes the most recent frame, but updates the memory slots every 5th frame. 
DAM4SAM is modified to update all RAM slots at every 5th frame (RAM$_{\mathrm{\overline{last}}}$). 
This results in a slight tracking quality decrease (1.3\%), which indicates that including the most recent frame in RAM is indeed beneficial, but not critical.

\subsubsection{Sensitivity analysis}\label{sec:sensitivity_analysis}

We analyze the sensitivity of the proposed DAM4SAM to the exact value of the manually determined parameters, defined in Section~\ref{sec:memory_management}. 
Experiments were conducted on the VOT2022~\citep{vot2022} dataset using the unsupervised experiment to ensure fast execution and compute the average overlap (AO) as the performance measure. 

Firstly, we examined the impact of different memory splits, specifically 2 frames in DRM and 4 frames in RAM, and vice versa. The results showed no significant changes, with performance drops of up to 0.3\%, indicating robustness to memory allocation variations.

We then evaluate the impact of varying the memory size in both RAM and DRM modules. Specifically, we experiment with configurations using fewer and more memory slots than the default. The tested settings include: 2 slots in both RAM and DRM, 4 slots in both RAM and DRM, and an asymmetric configuration with 6 slots in RAM and 3 slots in DRM. The results, measured by Average Overlap (AO) on the VOT2022-ST benchmark, are summarized in Table~\ref{tab:memory_ablation}. The default configuration with 3 slots in both RAM and DRM achieves the highest performance.

\begin{table}[th]
\centering
\caption{DAM4SAM memory size ablation on VOT2022 dataset.}
\begin{tabular}{ccl}
\toprule
 \# RAM slots & \# DRM slots & AO (\%) \\
\midrule
 2 & 2 & 0.781 \\
\textbf{3} & \textbf{3} & \textbf{0.792} \\
4 & 4 & 0.749 \\
6 & 3 & 0.759 \\
\bottomrule
\label{tab:memory_ablation}
\end{tabular}
\end{table}

\begin{figure}[ht]
  \centering
  \includegraphics[width=0.8\linewidth]{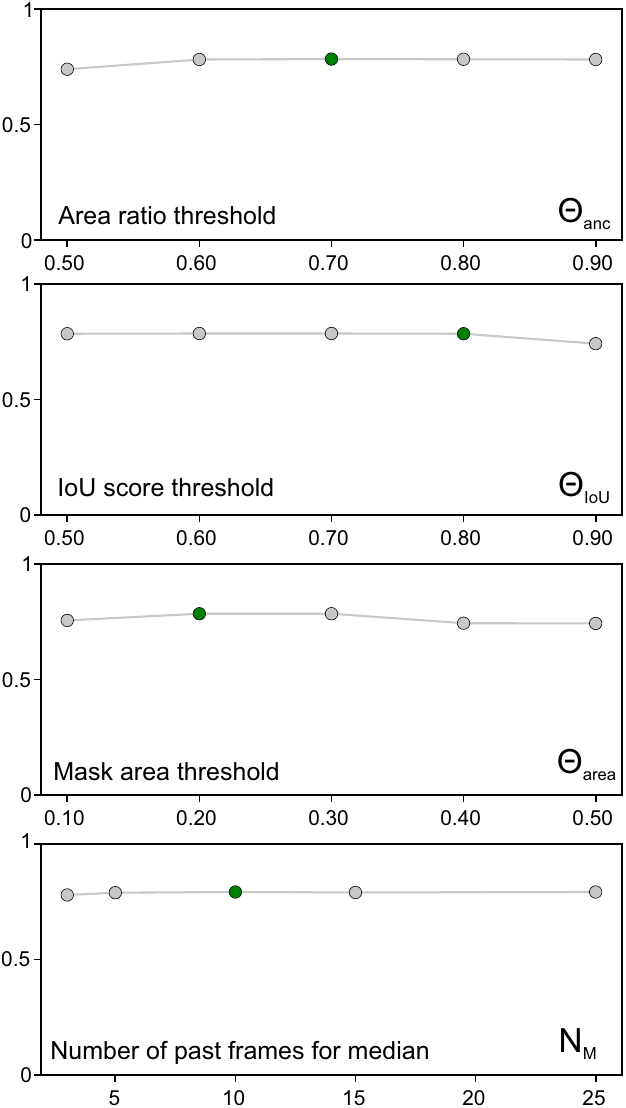} 
  \caption{
  Sensitivity of the DAM4SAM to different values of parameters: $\Theta_{anc}$, $\Theta_{IoU}$ and $\Theta_{area}$ and $N_{M}$. 
  Experiments were done on VOT2022 using average overlap as the performance measure. The selected parameter value is marked by a green circle. 
  }
\label{fig:suppl-parameters}
\end{figure}

Next, we analyzed the influence of the threshold for the ratio between the target and alternative predicted masks, i.e. $\Theta_{anc}$ = 0.7. 
This ratio is used for preemptive update upon distractor detection  and is thus crucial for our distractor-aware memory. 
As shown in Figure~\ref{fig:suppl-parameters}, tracking performance remains highly stable for a wide range of $\Theta_{anc} \in [0.6, 0.9]$ demonstrating the robust design of the tracker. 

Similarly, we evaluated the IoU score threshold ($\Theta_{Iou}$ = 0.8), which determines if a predicted mask is reliable for distractor testing. 
Results in Figure~\ref{fig:suppl-parameters} show that tracking performance remains consistent for a wide range of parameters ($\Theta_{IoU} \in [0.5, 0.8]$) scoring almost identical AO. 

The mask area threshold was also tested, i.e. $\Theta_{area}$ = 0.2.  This threshold is used to determine the tracking stability and is together with the $\Theta_{IoU}$ a necessary condition to trigger the DRM update. Tracking performance remained stable across variations of this parameter.

Finally, we evaluated the impact of the window size used in the median calculation for temporally consistent predictions, starting with an initial value of $N_{M} = 10$ and exploring a range of values three times larger and smaller. The performance variations were up to 1.5\% AO, indicating stability. 

Sensitivity analysis for parameters $\Theta_{anc}$, $\Theta_{IoU}$, $\Theta_{area}$ and $N_{M}$ across a wide range of thresholds is summarized in Figure~\ref{fig:suppl-parameters}. DAM4SAM demonstrates stable tracking performance, confirming that it is not sensitive to the exact value of these parameters.

\subsection{Comparison on DiDi}  \label{sec:exp-disd}

DAM4SAM is evaluated on the DiDi dataset against the baseline SAM2~\citep{sam2}, as well as several recent state-of-the-art trackers, including TransT~\citep{transt_cvpr2021}, SeqTrack~\citep{seqtrack}, AQATrack~\citep{aqatrack}, and AOT~\citep{ms_aot}. We also compare with trackers that incorporate explicit distractor handling mechanisms, such as KeepTrack~\citep{keeptrack}, Cutie~\citep{cutie_cvpr2024}, and ODTrack~\citep{odtrack}. Additionally, we include the latest, yet unpublished, SAM2 extensions: SAM2Long~\citep{sam2long}, which improves long-term tracking via memory tree, and SAMURAI~\citep{samurai}, which introduces motion-aware memory.

Results in Table~\ref{tab:dida} reveal the advantages of the trackers with an explicit distractor-handling mechanism over the other trackers. Consider two similarly complex recent trackers SeqTrack and ODTrack, which are both based on the ViT-L backbone. 
On classical benchmarks like LaSoT, LaSoT$_{\mathrm{ext}}$ and GoT10k, ODTrack outperforms SeqTrack by 2\%, 6\% and 4\%, respectively (see Table~\ref{tab:bbox}).
However, the performance gap increases to 15\% on DiDi (Table~\ref{tab:dida}), which confirms that distractors are indeed a major challenge for modern trackers and that DiDi has a unique ability to emphasize the tracking capability under these conditions and to expose tracking design weaknesses.

Focusing on the evaluation of the proposed tracker, DAM4SAM outperforms all trackers -- the standard state-of-the-art trackers as well as trackers with explicit distractor handling mechanism. 
In particular, DAM4SAM outperforms state-of-the-art distractor-aware trackers ODTrack and Cutie,  by 14\% and 21\%, respectively.

We compare the proposed DAM4SAM with the concurrent work SAMURAI~\citep{samurai}. Results show that DAM4SAM outperforms SAMURAI in tracking quality by 2\% on DiDi, mostly due to the higher robustness (i.e., DAM4SAM tracks longer than SAMURAI). 
This result demonstrates the superiority of our new DAM memory and the management protocol for distractor handling, which is also less complex than the concurent counterpart proposed in SAMURAI.

Compared to another unpublished tracker with the alternative memory design SAM2.1Long~\citep{sam2long}, DAM4SAM outperforms it by a 7\% tracking quality boost, indicating superiority of our proposed memory. The results reveal that the major source of the performance boost is the DAM4SAM tracking robustness, which means it fails less often and thus much better handles distractors.
In fact, a close inspection of the results shows that SAM2.1Long performs on par with the baseline SAM2.1, which indicates that the long-term memory update mechanism of SAM2.1Long does not improve performance in the presence of distractors. 
Finally, comparing DAM4SAM to the baseline SAM2.1 reveals 7\% boost in tracking quality, again, attributed mostly due to the improved robustness (6\%). These results validate the benefits of the proposed DAM and its management protocol in handling challenging distractors. 

\begin{table}[th]
  \centering
  \resizebox{\linewidth}{!}{\begin{minipage}{\linewidth}
\centering
\caption{State-of-the-art comparison on DiDi dataset.}
\label{tab:dida}
\begin{tabular}{llll}
\toprule
  & Quality & Acc. & Rob. \\
\midrule
SAMURAI \tiny{(arXiv24)} & 0.680 \second{} & 0.722 \third{} & 0.930 \second{} \\
SAM2.1Long \tiny{(arXiv24)} & 0.646 & 0.719 & 0.883  \\
ODTrack \tiny{(AAAI24)} & 0.608 & {\bf 0.740} \first{} & 0.809  \\
Cutie \tiny{(CVPR24)} & 0.575 & 0.704 & 0.776  \\
AOT \tiny{(NeurIPS21)} & 0.541 & 0.622 & 0.852  \\
AQATrack \tiny{(CVPR24)} & 0.535 & 0.693 & 0.753  \\
SeqTrack \tiny{(CVPR23)}  & 0.529 & 0.714 & 0.718  \\
KeepTrack \tiny{(ICCV21)} & 0.502 & 0.646 & 0.748  \\
TransT \tiny{(CVPR21)} & 0.465 & 0.669 & 0.678  \\
\midrule
SAM2.1 \tiny{(ICLR25)} & 0.649 \third{} & 0.720 & 0.887 \third{} \\
DAM4SAM & {\bf 0.694} \first{} & 0.727 \second{} & {\bf 0.944} \first{} \\
\bottomrule
\end{tabular}
\end{minipage}}
\end{table}

\subsection{Comparison on VOT challenges}  \label{sec:exp-vot}

The VOT initiative~\citep{kristan_vot_tpami2016, vot_website} is the major tracking initiative, providing challenging datasets for their yearly challenges. 
In contrast to most of the tracking benchmarks, the targets are annotated by segmentation masks, which allows more accurate evaluation of segmentation trackers, compared to the classic bounding-box benchmarks. 
In this paper, we include two recent single-target challenges VOT2020~\citep{kristan_vot2020} and VOT2022~\citep{vot2022} and the most recent multi-target challenge VOTS2024~\citep{vots2024}. 
We further evaluate the proposed DAM4SAM under real-time tracking constraints on VOT2022-RT~\citep{vot2022} challenge\footnote{SAM2.1 and DAM4SAM were evaluated on the machine with the AMD EPYC 7763 64-Core 2.45 GHz CPU and Nvidia A100 40GB GPU.}.

{\bf VOT2020 benchmark}~\citep{kristan_vot2020} consists of 60 challenging sequences, while trackers are run using anchor-based protocol~\citep{kristan_vot2020} to maximally utilize each sequence. 
Tracking performance is measured by the accuracy and robustness, summarized by the primary measure called the expected average overlap (EAO).

Results on VOT2020 are shown in Table~\ref{tab:vot20}. The proposed DAM4SAM outperforms all compared trackers. 
In particular, it outperforms the recently published MixViT~\citep{mixvit_tpami2024} by 25\% EAO, improving both, accuracy and robustness. 
DAM4SAM outperforms also the VOT2020 challenge winner RPT~\citep{rpt_eccvw2020} by a large margin (37.5\% in EAO). 
Comparing DAM4SAM to the original SAM2.1, the EAO is boosted by 7\%, while accuracy and robustness are improved by 2.7\% and 2.1\%, respectively. 

\begin{table}[th]
\centering
\resizebox{\linewidth}{!}{\begin{minipage}{\linewidth}
\centering
\caption{State-of-the-art comparison on the VOT2020 benchmark. The challenge winner is marked by~\includegraphics[height=0.7em]{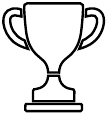}.
}
\label{tab:vot20}
\begin{tabular}{llll}
\toprule
  & EAO & Acc. & Rob. \\
\midrule
ODTrack \tiny{(AAAI24)} & 0.605 \third{} & 0.761 & 0.902 \third{} \\
MixViT \tiny{(TPAMI24)}  & 0.584 & 0.755 & 0.890  \\
SeqTrack \tiny{(CVPR23)} & 0.561 & - & -  \\
MixFormer \tiny{(CVPR22)} & 0.555 & 0.762 \third{} & 0.855  \\
RPT~\includegraphics[height=0.7em]{trophy_e.pdf} \tiny{(ECCVW20)}  & 0.530 & 0.700 & 0.869  \\
OceanPlus \tiny{(ECCV20)} & 0.491 & 0.685 & 0.842  \\
AlphaRef \tiny{(CVPR21)} & 0.482 & 0.754 & 0.777  \\
AFOD  \tiny{(ECCV20)} & 0.472 & 0.713 & 0.795  \\
\midrule
SAM2.1 \tiny{(ICLR25)} & 0.681 \second{} & 0.778 \second{} & 0.941 \second{} \\
DAM4SAM   & {\bf 0.729} \first{} & {\bf 0.799} \first{} & {\bf 0.961} \first{} \\
\bottomrule
\end{tabular}
\end{minipage}}
\end{table}

{\bf VOT2022 benchmark}~\citep{vot2022} uses a refreshed dataset with 62 sequences (simplest sequences removed, and more challenging added).  
Table~\ref{tab:vot22} includes the challenge top-performers, including the winner MS\_AOT~\citep{ms_aot}, as well as the recent published state-of-the-art trackers: DiffusionTrack~\citep{diffusiontrack}, MixFormer~\citep{mixformer_cvpr2022}, OSTrack~\citep{ostrack_eccv2022} and D3Sv2~\citep{d3s_tpami}. 
The proposed DAM4SAM outperforms the VOT2022 winner MS\_AOT by a significant margin of 12\% in EAO. 
Note that the performance improvement is a consequence of both improved accuracy (2\%) and robustness (3\%) compared to the MS\_AOT~\citep{ms_aot}. 
In addition to achieving state-of-the-art performance, DAM4SAM outperforms also the baseline SAM2.1 by a significant margin of 9\% EAO. 

The results on both, VOT2020 and VOT2022 clearly show that DAM4SAM
outperforms all trackers, including the challenges top-performers and the recently published trackers, setting new state-of-the-art on these benchmarks. 
Despite its simplicity, the proposed {\it training-free} memory management is the key element for achieving excellent tracking performance.

\begin{table}[th]
\centering
\resizebox{\linewidth}{!}{\begin{minipage}{\linewidth}
\centering
\caption{State-of-the-art comparison on the VOT2022 benchmark. The challenge winner is marked by~\includegraphics[height=0.7em]{trophy_e.pdf}.
}
\label{tab:vot22}
\begin{tabular}{llll}
\toprule
  & EAO & Acc. & Rob. \\
\midrule
MS\_AOT
~\includegraphics[height=0.7em]{trophy_e.pdf}  \tiny{(NeurIPS21)} & 0.673 \third{} & 0.781 & 0.944 \third{}  \\
DiffusionTrack \tiny{(AAAI24)} & 0.634 & - & -  \\
DAMTMask \tiny{(ECCVW22)} & 0.624 & 0.796 \third{} & 0.891  \\
MixFormer \tiny{(CVPR22)} & 0.589 & 0.799 \second{} & 0.878  \\
OSTrack \tiny{(ECCV22)} & 0.581 & 0.775 & 0.867  \\
Linker \tiny{(TMM24)} & 0.559 & 0.772 & 0.861  \\
SRATransT \tiny{(ECCVW22)} & 0.547 & 0.743 & 0.866  \\
TransT \tiny{(CVPR21)}  & 0.542 & 0.743 & 0.865  \\
GDFormer \tiny{(ECCVW22)} & 0.538 & 0.744 & 0.861  \\
TransLL \tiny{(ECCVW22)} & 0.530 & 0.735 & 0.861  \\
LWL\_B2S \tiny{(ECCVW20)} & 0.516 & 0.736 & 0.831  \\
D3Sv2 \tiny{(TPAMI21)} & 0.497 & 0.713 & 0.827  \\
\midrule
SAM2.1 \tiny{(ICLR25)} &  0.692 \second{} & 0.779 & 0.946 \second{} \\
DAM4SAM       & {\bf 0.753} \first{} & {\bf 0.800} \first{} & {\bf 0.969} \first{} \\
\bottomrule
\end{tabular}
\end{minipage}}
\end{table}

{\bf VOTS2024 benchmark}~\citep{vots2024} is one of the most recent tracking benchmarks, consisting of 144 sequences. 
In contrast to VOT2020 and VOT2022, the VOTS2024 benchmark introduces a new, larger dataset, tracking multiple objects in the same scene (with ground truth sequestered at an evaluation server) and a new performance measure, designed to address short-term, long-term\footnote{The target may leave the field of view and re-appear later in the video.}, single and multi-target tracking scenarios. 
The VOTS2024 is currently considered as the most challenging tracking benchmark. 

Although each sequence might contain several targets to track, the benchmark allows either to track these individually or jointly. The VOTS performance measures do not distinguish between how the tracks of all required targets are obtained, thus generalizing over both tracking approaches. As DAM4SAM is a single-target tracker, the targets are tracked individually in the VOTS2024 benchmark.

Results are reported in Table~\ref{tab:vot24}.
It is worth pointing out that the top performers are mostly unpublished (not peer-reviewed) trackers, tuned for the competition, and often complex ad-hoc combinations of multiple methods. 
For example, the challenge winner S3-Track combines visual and (mono)depth features, uses several huge backbones, and is much more complex than DAM4SAM.
Despite this, using the same parameters as in other experiments, DAM4SAM achieves a solid second place among the challenging VOTS2024 competition.
In particular, it outperforms the challenge-tuned versions of the recently published trackers LORAT~\citep{lorat_eccv2024}, Cutie~\citep{cutie_cvpr2024}, and the VOT2022 winner AOT~\citep{ms_aot}. 
Furthermore, the proposed memory management mechanism in DAM4SAM contributes to 8\% performance boost in tracking quality compared to the baseline SAM2.1, mostly due to 9\% higher robustness. 

\begin{table}[th]
  \centering
  \resizebox{\linewidth}{!}{\begin{minipage}{\linewidth}
\centering
\caption{State-of-the-art comparison on the VOT2024 benchmark. The challenge winner is marked by~\includegraphics[height=0.7em]{trophy_e.pdf}.}
\label{tab:vot24}
\begin{tabular}{llll}
\toprule
  & Quality & Acc. & Rob. \\
\midrule
S3-Track~\includegraphics[height=0.7em]{trophy_e.pdf} \tiny{(arXiv23)} & {\bf 0.722} \first{} & 0.784 & {\bf 0.889} \first{}  \\
DMAOT\_SAM \tiny{(NeurIPS22)} & 0.653 & {\bf 0.794}\first{} & 0.780  \\
HQ-DMAOT \tiny{(NeurIPS22)} & 0.639 & 0.754 & 0.790  \\
DMAOT \tiny{(NeurIPS22)} & 0.636 & 0.751 & 0.795 \third{}  \\
LY-SAM \tiny{(ECCVW24)} & 0.631 & 0.765 & 0.776  \\
Cutie \tiny{(CVPR24)} & 0.607 & 0.756 & 0.730  \\
AOT  \tiny{(NeurIPS21)}  & 0.550 & 0.698 & 0.767  \\
LORAT \tiny{(ECCV24)} & 0.536 & 0.725 & 0.784  \\
\midrule
SAM2.1 \tiny{(ICLR25)} & 0.661 \third{} & 0.791 \third{} & 0.790 \\
DAM4SAM  & 0.711 \second{} & 0.793 \second{} & 0.864 \second{} \\
\bottomrule
\end{tabular}
\end{minipage}}
\end{table}

\textbf{VOT2022-RT benchmark}~\citep{vot2022} was a challenge specifically designed for real-time evaluation. The sequences remain the same as in previously mentioned VOT2022. Specifically, VOT challenges are run by VOT toolkits, which manage the real-time constraints. A frame is sent to the tracker, which needs to process it and report the target position at 20FPS frame rate.
If the tracker is not able to process the frame in time, the prediction from the previous frame is used as the estimate for the current frame and next frame is sent to the tracker. 
Such a setup simulates actual real-time scenarios, which is much more realistic than reporting just the average tracking speed. The tracking performance is measured using standard VOT2022 measures~\citep{vot2022}: the primary measure expected average overlap (EAO), and two auxiliary measures, i.e., accuracy and robustness. 

The results in Table~\ref{tab:real_time} show that the proposed DAM4SAM (L model size) outperforms all trackers that participated in VOT2022-RT challenge. 
In particular, it outperforms the challenge winner by 4\% in EAO demonstrating excellent real-time performance. 
These results show that the proposed distractor-aware memory adds only a small computational overhead, yet bringing remarkable robustness capabilities and making it useful for real applications.

\begin{table}[th]
\centering
\resizebox{\linewidth}{!}{\begin{minipage}{\linewidth}
\centering
\caption{Real-time performance on VOT2022-RT challenge. The challenge winner is marked by~\includegraphics[height=0.7em]{trophy_e.pdf}.}
\label{tab:real_time}
\begin{tabular}{llll}
\toprule
  & EAO & Acc. & Rob. \\
\midrule
MS\_AOT~\includegraphics[height=0.7em]{trophy_e.pdf} \tiny{(NeurIPS21)}   & 0.610 \third{} & 0.751 \second{} & 0.921 \third{}  \\
OSTrack \tiny{(ECCV22)} & 0.569 & \textbf{0.766} \first{} & 0.860  \\
SRATransT \tiny{(ECCVW22)} & 0.547 & 0.743 \third{} & 0.866  \\
\midrule
SAM2.1 \tiny{(ICLR25)}   & 0.614 \second{} & 0.722 & 0.922 \second{} \\
DAM4SAM  & \textbf{0.635} \first{} & 0.717 & \textbf{0.942} \first{} \\
\bottomrule
\end{tabular}
\end{minipage}}
\end{table}

\subsection{Comparison on box datasets}  \label{sec:exp-bb}

For a complete evaluation, we compare DAM4SAM on the following three standard bounding box tracking datasets: LaSoT~\citep{lasot_cvpr19}, LaSoT$_{\mathrm{ext}}$~\citep{lasot_ijcv} and GoT10k~\citep{got10k}. 
Since frames are annotated by bounding boxes and SAM2 requires a segmentation mask provided in the first frame, we use the same SAM2 model to estimate the initialization mask. 
The min-max operation is applied on the predicted masks to obtain the axis-aligned bounding boxes required for the evaluation. 
Tracking performance is computed using area under the success rate curve~\citep{otb_pami2015} (AUC) in LaSoT~\citep{lasot_cvpr19} and LaSoT$_{\mathrm{ext}}$~\citep{lasot_ijcv} and the average overlap~\citep{got10k} (AO) in Got10k. 

{\bf LaSoT}~\citep{lasot_cvpr19} is a large-scale tracking dataset with 1,400 video sequences, with 280 evaluation sequences and the rest are used for training. 
The sequences are equally split into 70 categories, where each category is represented by 20 sequences (16 for training and 4 for evaluation). 
The dataset consists of various scenarios covering short-term and long-term tracking.
Results are shown in Table~\ref{tab:bbox}. 
The proposed DAM4SAM outperforms the baseline SAM2.1~\citep{sam2} by 7.3\%, which indicates that the proposed memory management is important also in a bounding box tracking setup. 
Furthermore, DAM4SAM performs on par with the top-performing tracker LORAT~\citep{lorat_eccv2024}, which was tuned on LaSoT training set, i.e., on the categories included in the evaluation set. 
It is worth noting that LORAT has approximately 50\% more training parameters than DAM4SAM, making the model significantly more complex. 

For additional insights on tracking performance, DAM4SAM is evaluated on fourteen visual attributes of the LaSoT dataset and compared with the baseline SAM2.1 in Table~\ref{tab:lasot-attributes}. 
The results show that the most challenging attributes are fast motion and full occlusion, which goes in line with the general findings of the past public attribute-based analyses~\citep{kristan_vot2019, kristan_vot2020}. 
Interestingly, DAM4SAM significantly outperforms SAM2.1 on these attributes, which demonstrates the effectiveness of the proposed design. 
On the other hand, the least challenging attributes are camera motion and motion blur, while the attributes for which the improvement is the smallest are partial occlusion, rotation and deformation. 
The results indicate that these attributes are already well addressed by the baseline method.

{\bf LaSoT$_{\mathrm{ext}}$}~\citep{lasot_ijcv} is an extension of the LaSoT~\citep{lasot_cvpr19} dataset, by 150 test sequences, divided into 15 new categories, which are not present in the training dataset. 
The results in Table~\ref{tab:bbox} show that DAM4SAM outperforms the baseline version by a comfortable margin of 7\% in AUC. 
In addition, it outperforms the second-best tracker LORAT~\citep{lorat_eccv2024} by 7.6\%. 
This indicates that DAM4SAM generalizes well across various object categories while existing trackers suffer a much larger performance drop.

{\bf GoT10k}~\citep{got10k} is another widely used large-scale tracking dataset, composed of $\sim$10k video sequences, from which 180 sequences are used for testing. 
We observed that top-performing trackers on GoT10k test set achieve excellent tracking performance, e.g., more than 78\% of average overlap, which leaves only small room for potential improvements. 
However, a solid 3.7\% boost in tracking performance is observed when comparing DAM4SAM to the top-performers LORAT~\citep{lorat_eccv2024} and ODTrack~\citep{odtrack}. 
A close inspection of the DAM4SAM results reveals that more than 99\% of frames are successfully tracked (i.e., with a non-zero overlap), which indicates that GoT10k ~\citep{got10k} difficulty level is diminishing for modern trackers.

\begin{table}[th]
\centering
\resizebox{\linewidth}{!}{\begin{minipage}{\linewidth}
\centering
\caption{State-of-the-art comparison on three standard bounding-box benchmarks: LaSoT, LaSoT$_{\mathrm{ext}}$, and GoT10k.}
\label{tab:bbox}
\begin{tabular}{llll}
\toprule
  & LaSoT & LaSoT$_{\mathrm{ext}}$ & GoT10k \\
  & (AUC) & (AUC) & (AO)  \\
\midrule
MixViT \tiny{(TPAMI24)} & 72.4 & -    & 75.7 \\
LORAT \tiny{(ECCV24)} & {\bf 75.1} \first{} & 56.6 \third{} & 78.2 \third{} \\
ODTrack \tiny{(AAAI24)} & 74.0 \second{} & 53.9 & 78.2 \third{} \\
DiffusionTrack \tiny{(AAAI24)} & 72.3 &  -  & 74.7 \\
DropTrack \tiny{(CVPR23)} & 71.8 & 52.7 & 75.9 \\
SeqTrack \tiny{(CVPR23)} & 72.5 \third{} & 50.7 & 74.8 \\
MixFormer \tiny{(CVPR22)} & 70.1 &  -  & 71.2 \\
GRM-256 \tiny{(CVPR23)} & 69.9 &  -   & 73.4 \\
ROMTrack \tiny{(ICCV23)} & 71.4 & 51.3 & 74.2 \\
OSTrack \tiny{(ECCV22)} & 71.1 & 50.5 & 73.7 \\
KeepTrack \tiny{(CVPR21)}  & 67.1 &  48.2   & -    \\
TOMP \tiny{(CVPR22)} & 68.5 &  -   & -    \\
\midrule
SAM2.1 \tiny{(ICLR25)} & 70.0 & 56.9 \second{} & 80.7 \second{} \\
DAM4SAM      & {\bf 75.1} \first{} & {\bf 60.9} \first{} & {\bf 81.1} \first{} \\
\bottomrule
\end{tabular}
\end{minipage}}
\end{table}

\begin{table*}[th]
\centering
\caption{Per-attribute comparison of SAM2.1 and DAM4SAM on LaSoT dataset. The AUC denotes overall performance on the whole dataset in terms of area-under-the-curve measure. The attribute abbreviations are the following: ARC~(aspect ratio change), BC~(background clutter), CM~(camera motion), DEF~(deformation), FM~(fast motion), FOC~(full occlusion), IV~(illumination variation), LR~(low resolution), MB~(motion blur), OV~(out-of-view), POC~(partial occlusion), ROT~(rotation), SV~(scale variation) and VC~(viewpoint change).}
\label{tab:lasot-attributes}
\scalebox{0.8}{
\begin{tabular}{llllllllllllllll}
\toprule
  & AUC & ARC & BC & CM & DEF & FM & FOC & IV & LR & MB & OV & POC & ROT & SV & VC  \\
\midrule
SAM2.1 \tiny{(ICLR25)}  & 70.0 & 68.8 & 65.3 & 70.7 & 71.7 & 60.2 & 60.6 & 65.9 & 61.5 & 69.2 & 63.3 & 69.3 & 68.2 & 69.5 & 62.9  \\
DAM4SAM                 & 75.1 & 74.1 & 70.8 & 79.8 & 76.1 & 68.3 & 68.7 & 75.1 & 69.2 & 77.2 & 70.2 & 73.4 & 72.2 & 74.8 & 73.6  \\
\midrule
Improvement (\%)             & 7.3 & 7.7 & 8.4 & 12.9 & 6.1 & 13.5 & 13.4 & 14 & 12.5 & 11.6 & 10.9 & 5.9 & 5.9 & 7.6 & 17  \\
\bottomrule
\end{tabular}
}
\end{table*}

Additionally, we evaluate the tracker's performance on three less commonly used benchmarks: OTB100~\citep{otb_pami2015}, NFS~\citep{nfs_iccv2017}, and UAV123~\citep{uav_benchmark_eccv2016}. The results are summarized in Table~\ref{tab:add_benchmarks}. Across all three benchmarks, DAM4SAM demonstrates a performance improvement over the baseline SAM2.1~\citep{sam2}, with gains ranging from 1.5\% to 3\%. On the OTB100~\citep{otb_cvpr2010} dataset specifically, DAM4SAM achieves a tracking robustness rate of 99.5\%, outperforming LORAT~\citep{lorat_eccv2024}, which reports 96.5\%.

\begin{table}[th]
\centering
\resizebox{\linewidth}{!}{\begin{minipage}{\linewidth}
\centering
\caption{State-of-the-art comparison on three additional bounding box benchmarks.}
\label{tab:add_benchmarks}
\begin{tabular}{llll}
\toprule
  & OTB100 & NFS & UAV123 \\
  & (AO) & (AUC) & (AUC)  \\
\midrule
MixFormer \tiny{(CVRP22)} & 70.7 \third{} & - & 70.4 \\
SeqTrack \tiny{(CVRP23)} & 68.3 & 66.2 & 68.5 \\
LORAT \tiny{(ECCV24)} & {\bf 72.0} \first{} & 66.7 \third{} &  \textbf{72.5} \first{} \\
\midrule
SAM2.1 \tiny{(ICLR25)} & 70.6 & 67.6 \second{} & 68.8 \third{} \\
DAM4SAM      & 71.7 \second{} & {\bf 68.6} \first{} & 70.9 \second{} \\
\bottomrule
\end{tabular}
\end{minipage}}
\end{table}

\subsection{Comparison on VOS datasets}  \label{sec:exp-vos}
To further assess the effectiveness of distractor-aware memory, we compare DAM4SAM with state-of-the-art methods across three video object segmentation benchmarks: VOST~\citep{vost_cvpr2023}, LVOS v1~\citep{lvos_iccv2023}, and LVOS v2~\citep{lvosv2_arxiv2024}. We report results on the semi-supervised VOS task. Methods are initialized with the ground truth segmentation mask and the predictions are evaluated with the $J\&F$ metric, an average of the VOS metrics \textit{J} (region similarity) and \textit{F} (contour accuracy).
Note that VOS benchmarks typically have a lower frame rate (5fps) compared to tracking datasets (30fps). 
To adapt the distractor-aware memory to this, we reduce the memory stride for both the RAM and DRM components from 5 to 1. This ensures that the memory updates created for tracking scenes align with the sparser annotation frequency in the VOS setting.

\textbf{VOST benchmark}~\citep{vost_cvpr2023} is designed for video object segmentation under complex object transformations. Unlike other datasets, VOST features objects that are broken, torn, or reshaped into entirely new forms, significantly altering their appearance. The validation set consists of 70 sequences, each averaging 21 seconds in length. Results are shown in Table~\ref{tab:exp_vos}. DAM4SAM outperforms the baseline SAM2.1~\citep{sam2} by 4.3\%, highlighting that the distractor-aware memory effectively enhances performance even in scenes where objects undergo significant transformations.

\textbf{LVOS datasets}~\citep{lvos_iccv2023, lvosv2_arxiv2024} is a long-term video object segmentation benchmarks. LVOS v2~\citep{lvosv2_arxiv2024} is an extended version of LVOS v1, increasing the number of sequences in the validation set from 50 to 140. On average, sequences span 68 seconds. The result in Table~\ref{tab:exp_vos} show that DAM4SAM outperforms the baseline SAM2.1~\citep{sam2} by 2.7\% and 6.1\%, on LVOS v1 and v2, respectively.

\begin{table}[th]
\centering
\resizebox{\linewidth}{!}{\begin{minipage}{\linewidth}
\centering
\caption{State-of-the-art comparison on video object segmentation benchmarks. Table reports $J\&F$ scores on validation sets.}
\label{tab:exp_vos}
\begin{tabular}{llll}
\toprule
  & VOST & LVOSv1  & LVOSv2 \\
\midrule
RDE \tiny{(CVPR22)} & -- & -- & 62.2 \\
DeAOT \tiny{(NeurIPS22)} & -- & -- & 63.9 \\
DEVA \tiny{(ICCV23)} & -- & 55.9 & -- \\
DDMemory \tiny{(ICCV23)} & --  & 60.7 & -- \\
XMem++ \tiny{(ICCV23)} & 37.5 & -- & 61.6 \\
Cutie \tiny{(CVPR24)} & 40.8 \third{} & 66.0 \third{} & 71.2 \third{} \\
\midrule
SAM2.1 \tiny{(ICLR25)} & 53.1 \second{} & 80.1 \second{} &  80.6 \second{} \\
DAM4SAM & \textbf{55.4} \first{} & \textbf{82.8} \first{} & \textbf{85.5} \first{} \\
\bottomrule
\end{tabular}
\end{minipage}}
\end{table}

\subsection{Qualitative analysis}\label{sec:qualitative}

\begin{figure}[ht]
  \centering
  \includegraphics[width=\linewidth]{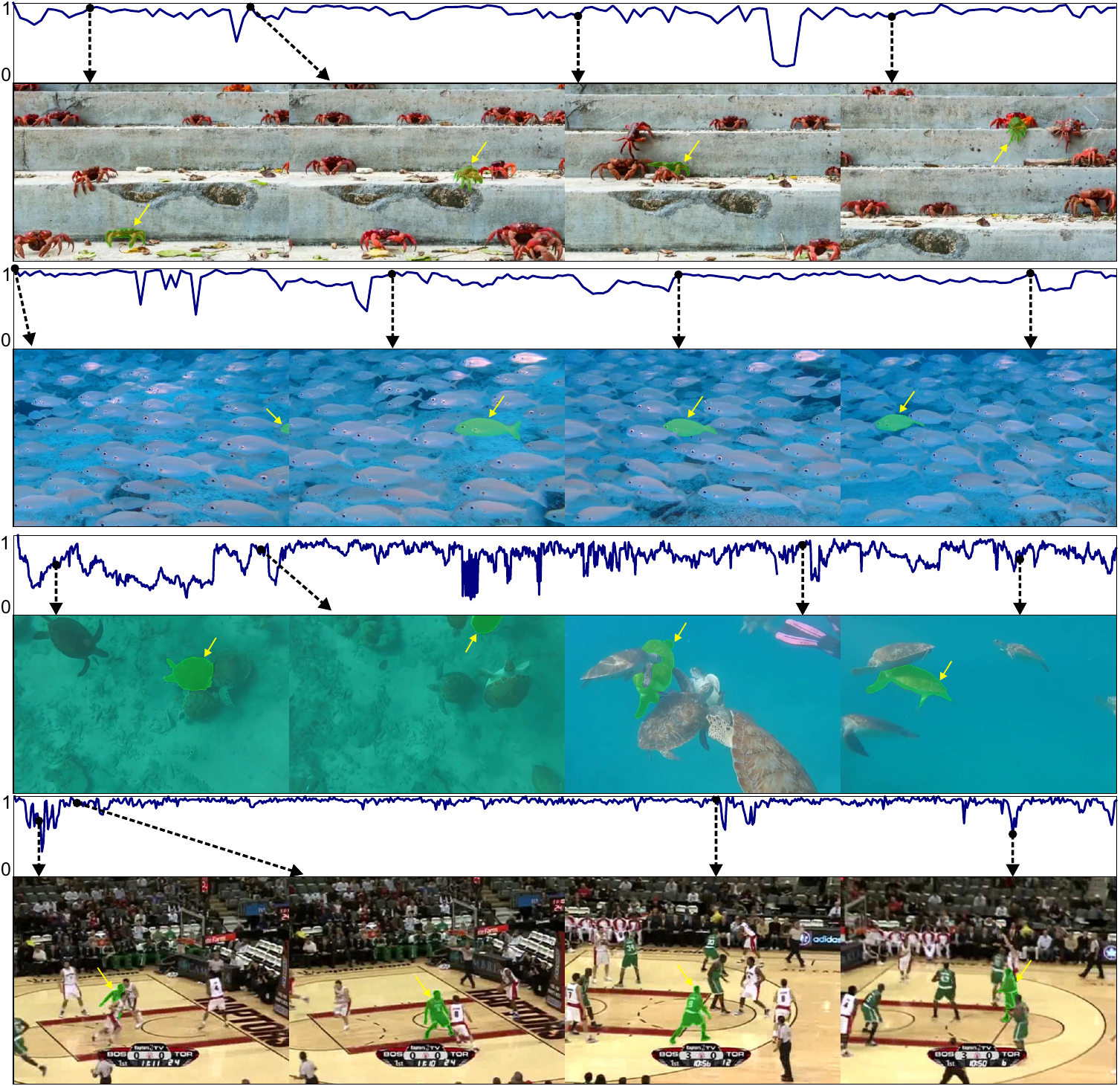} 
  \caption{DAM4SAM qualitative results on the DiDi dataset with predicted masks shown in green, and tracked objects denoted by arrows. 
  Per-frame overlaps (intersection-over-union for each frame) are shown above the figures to indicate failure-free tracking over the entire sequence.}
\label{fig:sam2pp_qualitative}
\end{figure}

Figure~\ref{fig:sam2pp_qualitative} presents qualitative results of DAM4SAM on several sequences from the DiDi dataset, illustrating its performance in the presence of challenging distractors. Across all sequences, DAM4SAM successfully tracks the target object without failure. It demonstrates strong robustness in crowded scenes, as seen in the first two examples. In the third sequence, where the task is to localize a turtle amidst multiple distractors and a background with similar appearance, DAM4SAM maintains accurate tracking. In the final example, taken from a basketball game, the model continues to perform reliably even with a fast-moving target.

\begin{figure*}[ht]
  \centering
  \includegraphics[width=\linewidth]{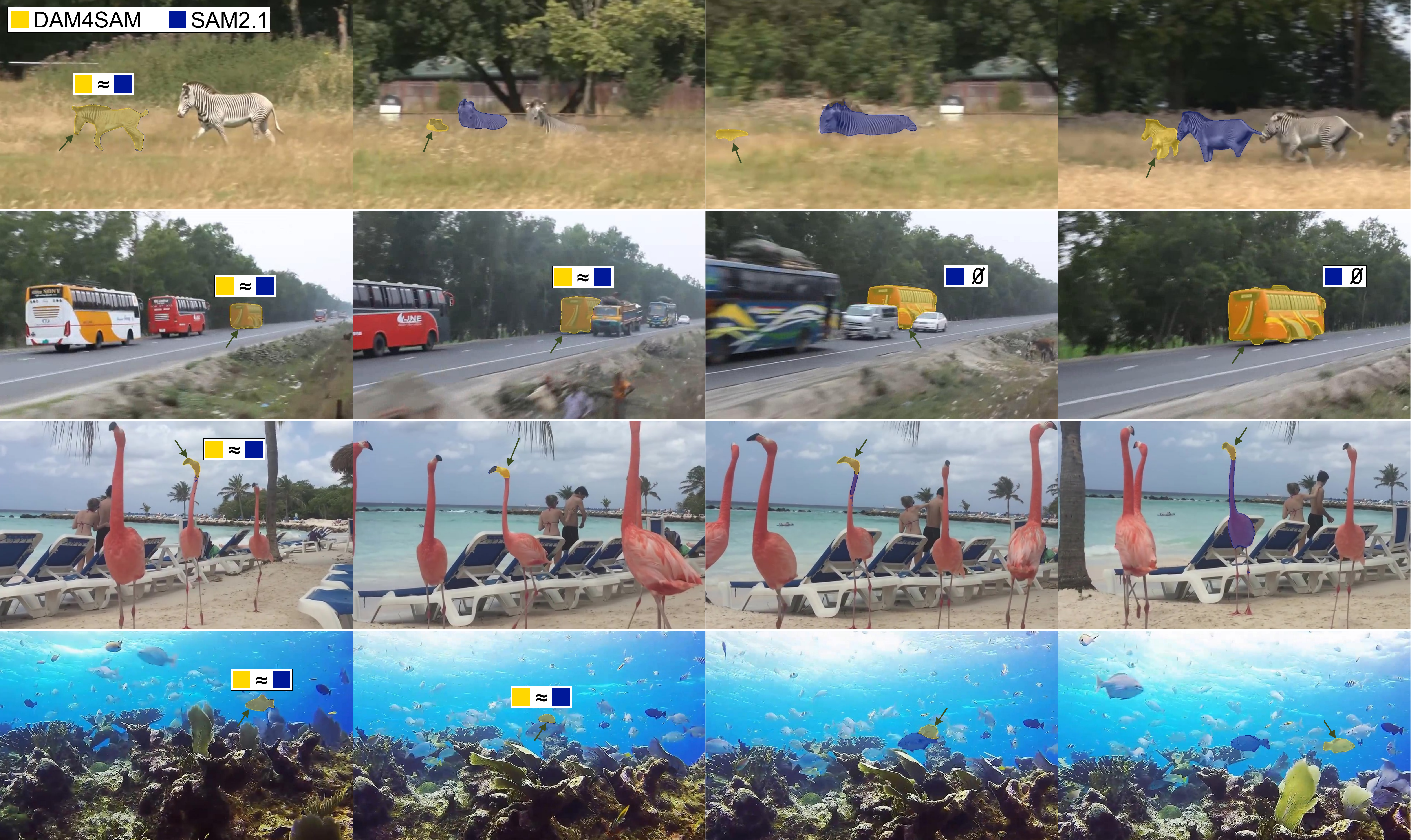} 
  \caption{Qualitative comparison of the baseline SAM2.1 (blue) and the proposed DAM4SAM (yellow). The symbol $\approx$ denotes approximately identical outputs and $\emptyset$ denotes an empty prediction (i.e, mask with all-zeros). Tracked object is denoted with a green arrow.}
\label{fig:qualitative_suppl}
\end{figure*}

Figure~\ref{fig:qualitative_suppl} presents a qualitative comparison between the baseline SAM2.1 and the proposed DAM4SAM on four video sequences from the DiDi dataset.
In the first row a zebra is tracked with other zebras in its vicinity. 
When the zebra is partially occluded, SAM2.1 drifts to the wrong zebra and starts to track it, while DAM4SAM tracks only the visible part of the target during occlusion and stays on the selected zebra until the end of the sequence.

In the second row of Figure~\ref{fig:qualitative_suppl}, the baseline SAM2.1 tracker successfully tracks the bus until the full occlusion and fails to re-detect it after the re-appearance. This failure occurs due to the too frequent memory updates when target is occluded and is successfully addressed with the proposed memory update in DAM4SAM.

The third row in Figure~\ref{fig:qualitative_suppl} shows tracking  of a flamingo's head. 
The baseline SAM2.1 tends to jump on the bird's beak or extend to the whole body, since it prefers to segment the regions with so-called high objectness (i.e., regions with well-defined edges). 
The proposed DAM4SAM successfully tracks the flamingo's head even if the edge between the head and the neck is not clearly visible. 
In this case, part of the neck is segmented by an alternative mask and thus detected as a distractor. 
Updating the distractor resolving memory (DRM) using such {\it critical} frames results in a more stable and accurate tracking.

A similar effect is demonstrated in the fourth row of Figure~\ref{fig:qualitative_suppl}, where a fish similar to the tracked fish occludes it and causes SAM2.1 to jump to it. 
On the other hand, DAM4SAM successfully detects such critical frames, updates the DRM and avoids the tracking failure.

\subsection{Failure cases}\label{sec:qa_failures}

\begin{figure*}[ht]
  \centering
    \includegraphics[width=\linewidth]{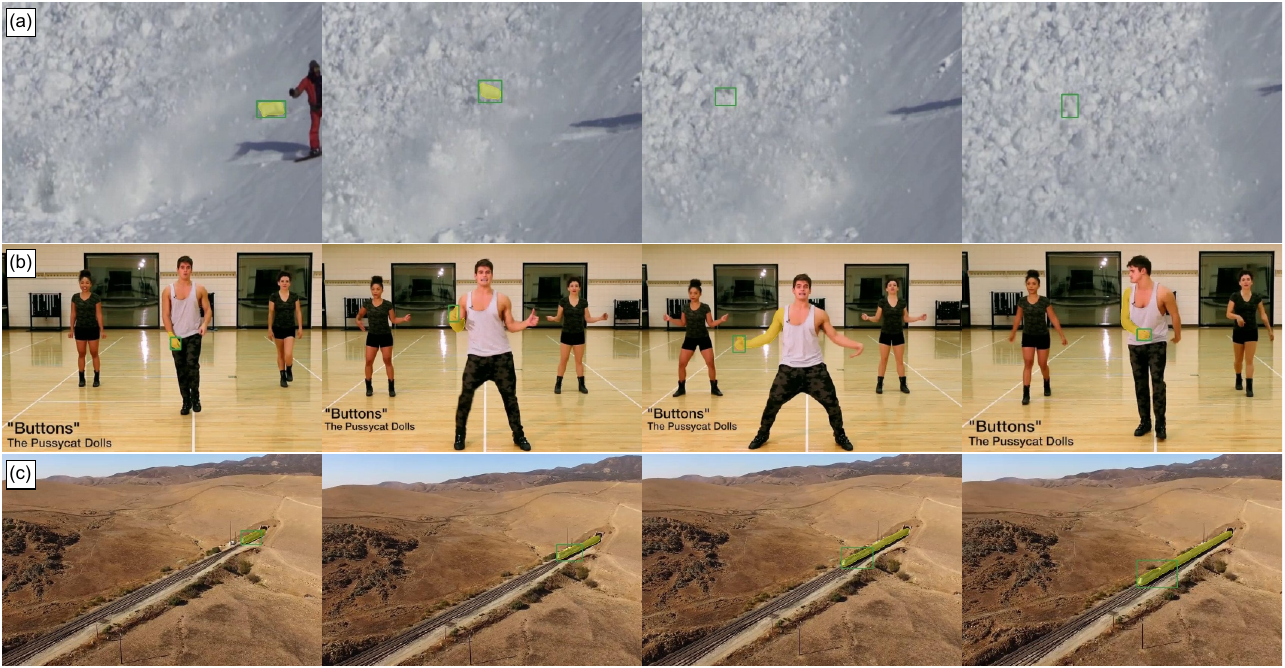} 
  \caption{Failure cases of DAM4SAM on DiDi dataset. The two most common reasons for failure are: (a) extreme visual similarity between the target and the background, and (b, c) overgeneralization to larger parts of the object when the goal is to track only a part of it. Yellow masks denote tracker predictions and green boxes are groundtruth.}
\label{fig:failure_cases}
\end{figure*}

An analysis of failure cases of the proposed tracker is presented in Figure~\ref{fig:failure_cases}. 
We identify two primary reasons why DAM4SAM fails. 
The first is the presence of extreme distractors, where the background closely resembles the target. One such example is shown in Figure~\ref{fig:failure_cases}(a). Although DAM4SAM successfully tracks the target through nearly half of the sequence, it eventually loses it due to the high visual similarity between the white rabbit (the target) and the avalanche in the background. It is worth noting that such cases are highly challenging even for human observers.

The second category of failure are generalizations to larger parts of the object, which we refer to as internal distractors. While these failures may significantly affect tracking accuracy, they do not compromise robustness. Examples are shown in Figure~\ref{fig:failure_cases}(b, c). 
A closer examination reveals that these issues may come from SAM2's limitation to generalize to large objects and its tendency to produce multiple plausible masks -- a behavior likely influenced by its training under uncertainty. 
We experimented with prompting SAM2 on the failure frame (where DAM4SAM overgeneralizes) using the first frame and its ground truth mask; the model would predict an overgeneralized mask, suggesting the error likely originates from SAM2 segmentation methodology itself.

\section{Generalization analysis} \label{sec:abl_studies}
In this section, we present experiments designed to investigate the generalization capabilities of the proposed drop-in memory. We begin by integrating DAM into different SAM2~\citep{sam2} model sizes and versions, presented in Section~\ref{sec:abl-model-size} and Section~\ref{sec:abl-model-version}, respectively.
To further demonstrate the effectiveness of distractor-aware memory when applied to a different tracker, Section~\ref{sec:dam4tam} presents results from integrating DAM into EfficientTAM~\citep{efficient_tam} and Section~\ref{sec:dam4edgetam} presents results when integrated into EdgeTAM~\citep{edgetam}. 

\subsection{Model size} \label{sec:abl-model-size}

The segment anything model 2 (SAM2)~\citep{sam2} was originally developed in four model sizes, denoted by tiny (T), small (S), base (B) and large (L). 
In Table~\ref{tab:model_size} these four model sizes of unchanged SAM2.1 are compared with our DAM4SAM version, presented in the paper on the new DiDi dataset. 
Results show a clear and consistent performance improvement across all four model sizes. 
In particular, the tracking quality improves by approximately  6\% or 7\%, depending on the model size, mostly due to the improved robustness. 
These results show 
that the proposed distractor-aware memory generalizes well over various model sizes, demonstrating that the model has not been tuned to the exact SAM2 model.

\begin{table}[h]
\centering
\resizebox{\linewidth}{!}{\begin{minipage}{\linewidth}
\centering
\caption{Comparison between SAM2.1 ({\it SAM}) and DAM4SAM ({\it D4S}) with different model sizes on DiDi dataset. The T, S, B, L denote the tiny, small, base and large Hiera backbone sizes, respectively. {\it Param} denotes number of parameters, {\it Mem} memory footprint during inference in gigabytes, {\it FPS} inference speed in frames-per-second, while {\it Acc.} and {\it Rob.} denote accuracy and robustness, respectively.}
\label{tab:model_size}
\begin{tabular}{lcccccc}
\toprule
  & Param & Mem & FPS & Quality & Acc. & Rob. \\
\midrule
SAM-T    & 39M  & 0.63 & 17.4 & 0.600 & 0.697 & 0.848 \\
D4S-T   & 39M  & 0.63 & 15.2 & 0.642 & 0.695 & 0.907 \\  \hdashline
SAM-S    & 46M  & 0.68 & 17.0 & 0.630 & 0.718 & 0.866 \\
D4S-S   & 46M  & 0.68 & 15.3 & 0.668 & 0.709 & 0.930 \\ \hdashline
SAM-B    & 81M  & 0.90 & 16.0 & 0.624 & 0.721 & 0.856 \\
D4S-B   & 81M  & 0.90 & 13.7 & 0.664 & 0.709 & 0.930 \\ \hdashline
SAM-L    & 224M & 1.78 & 13.2 & 0.649 & 0.720 & 0.887 \\
D4S-L   & 224M & 1.78 & 11.8 & 0.694 & 0.727 & 0.944 \\
\bottomrule
\end{tabular}
\end{minipage}}
\end{table}

\subsection{Model version}  \label{sec:abl-model-version}

This section compares the performance improvements for the two individual SAM versions, i.e., SAM2 and SAM2.1. The SAM2.1 version improves the initial version in handling small and visually similar objects by introducing additional augmentation techniques in training. 
It also includes improved occlusion handling by training the model on longer frame sequences. 

Results are shown in  Table~\ref{tab:model_version}.
To demonstrate the improvement of the 2.1 model version over version 2, we compare the SAM2 with the SAM2.1 on the DiDi dataset, which results in approximately 3.5\% improvement in tracking quality. 
Next, we compare the original SAM2 and the version with our new memory model. The tracking performance improves by 7\%, which is well beyond the performance improvement from SAM2 to SAM2.1 and supports the importance of a high-quality memory model and the memory management regime.
A similar performance boost (close to 7\%) is observed between SAM2.1 and DAM4SAM, which implies complementarity of the new memory model with the baseline method performance imporvements that come from better training.
Similarly as in Section~\ref{sec:abl-model-size}, we conclude that the proposed distractor-aware memory is robust to different model versions, demonstrating a consistent improvements in tracking performance on two SAM2 versions. 

\begin{table}[h]
\centering
\caption{Comparison of SAM2 and SAM2.1 on DiDi dataset.}
\label{tab:model_version}
\begin{tabular}{llll}
\toprule
  & Quality & Acc. & Rob. \\
\midrule
SAM2 \tiny{(ICLR25)}     & 0.627 & 0.723 & 0.850 \\
DAM4SAM (2)    & 0.668 $\uparrow$7\% & 0.710 & 0.929 \\
\hdashline
SAM2.1 \tiny{(ICLR25)}   & 0.649 & 0.720 & 0.887 \\
DAM4SAM (2.1) & 0.694 $\uparrow$7\% & 0.727 & 0.944 \\
\bottomrule
\end{tabular}
\end{table}

\subsection{Lightweight architecture} \label{sec:dam4tam}

To further validate the advantages of the proposed distractor-aware memory, we integrate it into a recent tracking model, EfficientTAM~\citep{efficient_tam}, an efficient variant of SAM2~\citep{sam2}. The authors propose using a lightweight image encoder to achieve approximately $2\times$ speedup and $2.4\times$ parameter reduction (34M compared to 81M), while maintaining performance comparable to the SAM2-B model. The memory mechanism follows the principles of SAM2, with the memory bank storing the last six frames along with the initialization frame, regardless of target visibility. 

The modified model, EfficientTAM with integrated DAM, is referred to as DAM4TAM. A performance comparison on the DiDi dataset is presented in Table~\ref{tab:dam4tam_didi_eval}. DAM4TAM achieves a 12\% improvement in overall quality over EfficientTAM on the DiDi dataset, due to a substantial increase in tracking robustness. Although initial comparisons between SAM2.1-S and EfficientTAM-S show a significant quality gap (with the latter performing 4.3\% worse), integrating DAM into both models reduces this gap to just 1.3\%. DAM4TAM runs at 25~fps, which is comparable to the speed of TAM-S.\footnote{All trackers were evaluated on a machine equipped with an AMD EPYC 7763 64-Core 2.45 GHz CPU and an NVIDIA A100 40GB GPU.}

\begin{table}[h]
\centering
\caption{Comparison of different versions of SAM2, EfficientTAM and EdgeTAM with the proposed DAM variants on DiDi.}
\label{tab:dam4tam_didi_eval}
\begin{tabular}{lllll}
\toprule
  & Q & Acc. & Rob. \\
\midrule
EfficientTAM-S \tiny{(arXiv24)} & 0.587 & 0.706 & 0.821 \\
DAM4TAM-S & 0.653 $\uparrow$11\% & 0.704 & 0.915 \\
\hdashline
EdgeTAM \tiny{(CVPR25)} & 0.566 & 0.673 & 0.821 \\
DAM4EdgeTAM & 0.591 $\uparrow$4\% & 0.669 & 0.858 \\
\hdashline
SAM2.1-S \tiny{(ICLR25)} & 0.630 & 0.718 & 0.866 \\
DAM4SAM-S & 0.668 $\uparrow$6\% & 0.709 & 0.930 \\ 
\hdashline
SAM2.1-L \tiny{(ICLR25)} & 0.649 & 0.720 & 0.887 \\
DAM4SAM-L & 0.694 $\uparrow$7\% & 0.727 & 0.944 \\
\bottomrule
\end{tabular}
\end{table}

DAM4TAM is evaluated against the baseline EfficientTAM~\citep{efficient_tam} on VOT2022 tracking dataset~\citep{vot2022} and two video object segmentation datasets: VOST~\citep{vost_cvpr2023} and LVOSv2~\citep{lvosv2_arxiv2024}. The results are presented in Table~\ref{tab:dam4tam_vots_eval}. On VOT2022, DAM4TAM improves baseline's performance by 8\%, mostly due to a robustness score of 0.938, while using 6.5$\times$ fewer parameters than SAM2.1-L (which has a robustness score of 0.946). On VOST, DAM4TAM outperforms EfficientTAM by 5\% in EAO, again achieving result comparable to SAM2.1-L.

\begin{table}[h]
\centering
\resizebox{\linewidth}{!}{\begin{minipage}{\linewidth}
\centering
\caption{Comparison of baseline and DAM4TAM on VOT2022, VOST, and LVOSv2 datasets.}
\label{tab:dam4tam_vots_eval}
\begin{tabular}{llll}
\toprule
  & VOT2022 & VOST & LVOSv2  \\
\midrule
EfficientTAM-S \tiny{(arXiv24)} & 0.585 & 0.509 & 0.805 \\
DAM4TAM-S & 0.630 $\uparrow$8\% & 0.537 $\uparrow$6\% & 0.817 $\uparrow$1.5\%\\
\bottomrule
\end{tabular}
\end{minipage}}
\end{table}

\subsection{Edge architecture} \label{sec:dam4edgetam} \label{sec:dam4edgetam}
Similar to EfficientTAM~\citep{efficient_tam}, a recent approach called 
EdgeTAM~\citep{edgetam} was proposed to enable SAM2~\citep{sam2}-based models to run on mobile devices while maintaining comparable performance. 
The authors identify that, in addition to the image encoder, the memory attention blocks are a major latency bottleneck. 
To address this, they introduce a 2D spatial perceiver module that significantly reduces computational cost. 
EdgeTAM runs with 15 FPS on an iPhone, compared to only 0.7 FPS for SAM2. 
For the memory management, EdgeTAM follows the same principles as SAM2.

The modified version with integrated DAM is referred to as DAM4EdgeTAM. 
Table~\ref{tab:dam4tam_didi_eval},  compares DAM4EdgeTAM to the baseline on the DiDi dataset. 
The DAM-enhanced model achieves improved performance, due to an approximate 4\% better robustness on the dataset.

\section{Conclusion}  \label{sec:conclusion}
We proposed a drop-in, training-free Distractor-Aware Memory (DAM) model and a corresponding memory management regime for memory-based trackers. The DAM architecture separates memory by tracking functionality into two components: Recent Appearances Memory (RAM) for maintaining segmentation accuracy, and Distractor-Resolving Memory (DRM) for enhancing robustness by supporting tracking recovery after failure and re-detection. We introduce efficient, introspection-based update rules for both memory types, relying solely on the tracker's outputs without additional training.

For fair performance evaluation in scenes with distractors, we construct DiDi, a distractor distilled dataset consisting of 180 video sequences with an average length of 1{,}500 frames. 
The dataset construction process (and manual inspection) ensure presence of distractors in each video, which makes DiDi a unique tracking benchmark. Since distractors are one of key tracking challenges, leading to tracking failures, DiDi has a strong potential to facilitate development of future distractor-robust trackers.

DAM is implemented on top of SAM2.1~\citep{sam2}, forming DAM4SAM. Extensive analysis and ablation studies support our design choices. 
Using frozen SAM2.1-L weights, DAM4SAM achieves state-of-the-art performance on ten benchmarks, with only a moderate 10\% speed reduction compared to SAM2.1. 
Tracking speed is not affected by the number of distractors in the scene.
On the other hand, more distractors in general may lead to more potential points of failure. 
Nevertheless, we did not observe a correlation between a number of distractors and low tracking performance.

Furthermore, DAM is integrated into multiple versions (v2 and v2.1) and sizes (tiny, small, base, and large) of the SAM2 model, consistently achieving a 6--7\% improvement on the DiDi dataset. 
For efficient tracking we integrated DAM into two recent SAM-based architectures: Efficient Track Anything~\citep{efficient_tam} and EdgeTAM~\citep{edgetam}.  
The DAM-enhanced versions consistently outperform the baseline on multiple tracking and segmentation benchmarks. 
These results indicate that DAM4SAM is practically useful in all standard tracking applications, such as surveillance, video editing and sports analytics, and extends to realtime and even edge-device tracking setups. 
Since it is able to provide precise object segmentation, not just an approximate object position as is the case for bounding box trackers, the application spectrum is potentially wider.

Failure case analysis of DAM4SAM reveals unresolved ambiguities in the foundational model SAM2~\citep{sam2}, regarding object boundaries. 
This issue likely comes from training on segmentation datasets, rather than tracking-specific data. 
SAM2 exhibits a tendency to favor objects with clear boundaries, which can lead to inconsistencies when tracking only part of an object. 

The results also suggest that more research should focus on efficient memory designs, with possibly learnable management policies. 
Such formulation, which we plan to explore in future work, would make the method even more. In fact, following the DAM4SAM conference publication~\citep{dam4sam}, preliminary works~\citep{sam2rl} emerged attempting to follow this avenue. 
Moreover, it is expected that the development of a better foundation model, which is integrated with DAM, would yield even better tracking performance.

\bmhead{Funding declaration}

This work was supported by Slovenian research agency program P2-0214 and projects 
J2-2506, 
Z2-4459,
and J2-60054, 
and by supercomputing network SLING (ARNES, EuroHPC Vega - IZUM).

\bibliography{sn-bibliography}

\end{document}